\documentclass[10pt,twocolumn,letterpaper]{article}

\usepackage{cvpr}
\usepackage{times}
\usepackage{epsfig}
\usepackage{graphicx}
\usepackage{amsmath}
\usepackage{amssymb}

\usepackage[utf8]{inputenc} %
\usepackage[T1]{fontenc}    %
\usepackage{url}            %
\usepackage{booktabs}       %
\usepackage{amsfonts}       %
\usepackage{nicefrac}       %
\usepackage{microtype}      %
\usepackage{epsfig}
\usepackage{graphicx}
\usepackage{amsmath}
\usepackage{amssymb}
\usepackage{booktabs}
\usepackage{wasysym}
\usepackage{xcolor} 
\usepackage{bbm}
\usepackage{multirow}
\usepackage{mathtools}
\usepackage{wrapfig}
\usepackage{cancel}
\usepackage{ulem}
\usepackage{caption}
\usepackage{subcaption}
\usepackage{authblk}

\usepackage{algorithm}
\usepackage{listings}

\usepackage[pagebackref=true,breaklinks=true,letterpaper=true,colorlinks=true,bookmarks=false]{hyperref}

\cvprfinalcopy %

\def\cvprPaperID{4896} %

\usepackage[font=small,labelfont=bf]{caption}

\ifcvprfinal\fi

\newif\ifcommentson

\ifcommentson
    
    \newcommand{\KG}[1]{{\color{blue}{}#1}}
    \newcommand{\todo}[1]{\textcolor{green}{\{todo: #1\}}}
    \newcommand{\KGnote}[1]{{\color{red}{}#1}}
    \newcommand{\comm}[1]{\textcolor{brown}{\{\small comment: #1\}}}
    \newcommand{\SPDelete}[1]{\sout{#1}}
    
\else
    
    \newcommand{\KG}[1]{#1}
    \newcommand{\todo}[1]{}
    \newcommand{\KGnote}[1]{}
    \newcommand{\comm}[1]{}
    \newcommand{\SPDelete}[1]{}
    
\fi

\usepackage{etoolbox}
\makeatletter
\AfterEndEnvironment{algorithm}{\let\@algcomment\relax}
\AtEndEnvironment{algorithm}{\kern2pt\hrule\relax\vskip3pt\@algcomment}
\let\@algcomment\relax
\newcommand\algcomment[1]{\def\@algcomment{\footnotesize#1}}
\renewcommand\fs@ruled{\def\@fs@cfont{\bfseries}\let\@fs@capt\floatc@ruled
  \def\@fs@pre{\hrule height.8pt depth0pt \kern2pt}%
  \def\@fs@post{}%
  \def\@fs@mid{\kern2pt\hrule\kern2pt}%
  \let\@fs@iftopcapt\iftrue}
\makeatother

\def\modelname{AV-Map }

\begin{document}

\title{Audio-Visual Floorplan Reconstruction}

\renewcommand\Authsep{~~~~ }
\renewcommand\Authands{~~~~ }
\author[1]{Senthil Purushwalkam\thanks{work done while interning at Facebook AI Research. Project webpage: \url{http://www.cs.cmu.edu/\~spurushw/publication/avmap}}}
\author[2]{Sebastian Vicenc Amengual Gari}
\author[2]{Vamsi Krishna Ithapu}
\author[2]{\\Carl Schissler}
\author[2]{Philip Robinson}
\author[3]{Abhinav Gupta}
\author[3,4]{Kristen Grauman}

\affil[1]{Carnegie Mellon University}
\affil[2]{Facebook Reality Labs}
\affil[3]{Facebook AI Research}
\affil[4]{University of Texas at Austin}
\maketitle

\begin{abstract}
Given only a few glimpses of an environment, how much can we infer about its entire floorplan?
Existing methods can map only what is visible or immediately apparent from context, and thus require substantial movements through a space to fully map it.  
We explore how both audio and visual sensing together can provide rapid floorplan reconstruction from limited viewpoints.  Audio not only helps sense geometry outside the camera's field of view, but it also reveals the existence of distant freespace (e.g., a dog barking in another room) and suggests the presence of rooms not visible to the camera (e.g., a dishwasher humming in what must be the kitchen to the left).  We introduce AV-Map, a novel multi-modal  encoder-decoder framework that reasons jointly about  audio and vision to reconstruct a floorplan from a short input video sequence.  We train our model to predict both the interior structure of the environment and the associated rooms' semantic labels. Our results on 85 large real-world environments show the impact: with just a few glimpses spanning 26\% of an area, we can estimate the whole area with 66\% accuracy --- substantially better than the state of the art approach for extrapolating visual maps.
\end{abstract}

\section{Introduction}
\label{sec:intro}

\begin{figure}[t]
    \centering
    \includegraphics[width=\linewidth]{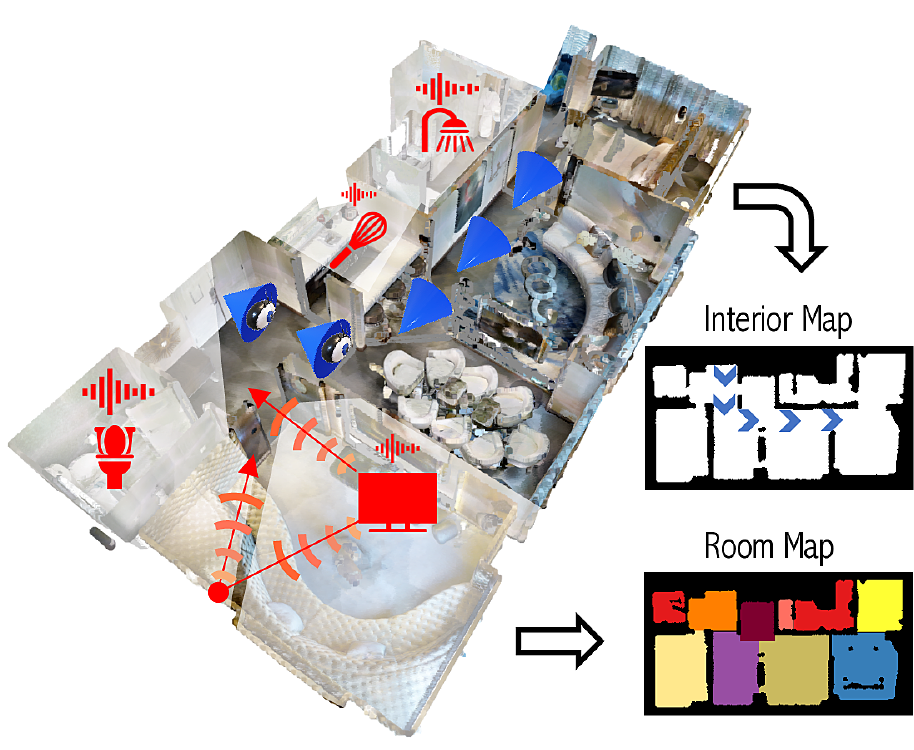}
    \vspace*{-0.2in}
    \caption{\textbf{Audio-visual floorplan reconstruction}: A short video walk through the house can reconstruct the visible portions of the floorplan, but is blind to many areas.
    We introduce \emph{audio-visual floorplan reconstruction}, where sounds in the environment help infer both the geometric properties of the hidden areas as well as the semantic labels of the unobserved rooms (e.g., sounds of a person cooking behind a wall to the camera's left suggest the kitchen). 
}\vspace*{-0.1in}
    \label{fig:teaser}
\end{figure}

Floorplans of complex 3D environments---such as homes, offices, shops, churches---are a compact ground-plane representation of their overall layout, showing the different rooms and their connectivity.  
Floorplans are useful for visualizing a large space, navigating an unfamiliar building, planning safety routes, and communicating architectural designs.  In robotics, an agent entering a new building needs to quickly sense the overall layout, but without visiting every part of it.

Traditionally a floorplan is created by distilling a \emph{fully observed} 3D environment into its footprint---whether manually or with the aid of 3D sensors~\cite{sui,okorn}.
Recent research aims to infer room layouts using imagery and/or scans, with impressive results~\cite{liu2018floornet,chen2019floor,roomnet,dulanet}.  However, existing methods are limited to mapping the regions they directly observe.
They either require a dense walk-through for the camera to capture most of the space---wasteful if not impossible for a robotic agent trying to immediately perform tasks in a new environment---or else they simply fail to map rooms beyond those where the camera was placed.  %

Our idea is to ``see" beyond the visible regions by listening.  
Audio provides strong spatial and semantic signals that complement the mapping capabilities of visual sensing. In particular, the value of audio for floorplan estimation is threefold.  First, observed sound is inherently driven by geometry; audio reflections bounce off major surfaces and reveal the shape of a room, beyond the camera's field of view.  Second, sounds heard from afar---even multiple rooms away---can suggest the existence of distant freespace where the sounding object could exist (e.g., a dog barking in another room).  Third, hearing semantically meaningful sounds from different directions naturally reveals the plausible room layouts %
based on the activities or objects those sounds represent.  For example, a shower running suggests the direction of the bathroom, even before we see it; microwave beeps suggest a kitchen; climbing footsteps suggest a staircase.  See Figure~\ref{fig:teaser}.

To this end, we propose a new research direction: \emph{audio-visual floorplan reconstruction}. %
Given a short RGB video complete with multi-channel audio, the goal is to produce a 2D floorplan that shows the freespace and occupied regions and divides them into  a discrete set of semantic room labels (family room, kitchen, etc.).  Importantly, the floorplan output extends significantly beyond the area directly observable in the video frames.  This efficiency is critical for navigating robots that need to act without exhaustively touring a space, as well as offline scenarios where a user wants to extract a broad map from an existing short video. %

Our \modelname approach works as follows.  We devise a deep convolutional neural network architecture that leverages sequences of audio and visual data to reason about the structure and semantics of the floorplan. Our model independently extracts floorplan-aligned features for audio and RGB data, %
encodes sequences of features of each modality using self-attention mechanisms, and finally fuses information from audio and RGB using a decoder architecture.  %

We consider two settings: device-generated sounds (active) and environment-generated sounds (passive).  In the active setting, the camera emits a known sound while it moves.  This corresponds to a use case where a person or robot does a swift walk-through of an environment while their phone/camera emits some sound.  In the passive setting, we observe only naturally occurring sounds made by objects and people in the building.  
This corresponds to a use case where we are simply given a passively recorded video, likely captured for some other purpose.

To our knowledge, ours is the first attempt to infer floorplans from audio-visual data.  Our results on 85 large real-world, \KG{multi-room} environments show that AV-Map not only consistently outperforms traditional vision-based mapping, but also improves the state-of-the-art approach~\cite{ramakrishnan2020occupancy} for extrapolating %
occupancy maps beyond visible regions \KG{(with a relative gain of 8\% in floorplan accuracy)}. %
Though observing only a small fraction of the full homes, our model yields good interior maps covering much of their area.
We also show audio and vision are \KG{synergistic} signals to classify room types, \KG{allowing high-level perception of the semantics of the space even before directly visiting each room.}

\section{Related Work}
\label{sec:related}

\paragraph{Floorplan and room layout reconstruction}

The vision and graphics communities have explored various ways to use visual data, depth sensors, and laser scanners to build floorplans.  Geometric approaches use 3D point cloud inputs to construct building-wide floor plans~\cite{sui,okorn}.  Given RGB-D scans, FloorNet~\cite{liu2018floornet} and Floor-SP~\cite{chen2019floor} estimate 2D floorplans and rooms' semantic labels using a mix of deep learning and optimization.  Given monocular RGB images~\cite{roomnet} or $360^\circ$ panoramas~\cite{dulanet,layoutnet,horizonnet,zou20193d}, other methods estimate a 3D indoor Manhattan room layout.  Using only a small portion of a $360^\circ$ panorama, models can be trained to infer missing viewpoints~\cite{look-around} and/or semantic labels~\cite{im2pano3d}.  Unlike any of the above, our approach leverages both audio and visual sensing to infer a 2D floorplan map and its semantic room labels. As our results show, audio offers the advantage of sensing further beyond the field of view of visual sensors.

\vspace*{-0.15in}
\paragraph{Mapping for navigation}
With adequate overlapping views, structure-from-motion %
methods can recover the 3D structure of an environment (e.g.,~\cite{lost,little}).  %
Laser-based 2D SLAM is often used in mobile robotics to obtain the ground plane map~\cite{2dslam}.
Recent work leverages scans of
indoor environments~\cite{chang2017matterport3d,straub2019replica} and fast simulation tools~\cite{savva2019habitat} %
to facilitate work on embodied visual %
navigation~\cite{gupta2017unifying,savinov2018semi,kadian2019we,chaplot2020learning,chen2019learning,ramakrishnan2020occupancy}. While often the map is implicitly learned, some  methods explicitly estimate a 2D occupancy map, projecting the observed point cloud to the ground plane and growing the map over time~\cite{chen2019learning,ramakrishnan2020occupancy,chaplot2020learning}.  To navigate to a specified room, the method of~\cite{narasimhan2020seeing} predicts a 2D semantic map with room labels, learning the layout patterns in houses.  In contrast to navigation, where an intelligent agent controls the camera and builds its map in service of reaching a target, our goal is to transform a passive video sequence (with audio) into a map.  We show the advantages of our audio-visual approach over OccAnt~\cite{ramakrishnan2020occupancy}, the state-of-the-art navigation method \KG{that} extrapolates beyond visible scene points using vision alone.

\vspace*{-0.15in}
\paragraph{Audio for spatial sensing}

Prior work explores ways to exploit audio alone to sense the shape of a room or object.  Given multiple microphone recordings of a known sound, the method of~\cite{dokmanic2013acoustic} computes the shape of a single convex polyhedral room, %
while sound reflections are used to sense the 3D shape of an object hidden around a corner~\cite{lindell2019acoustic}.  In robotics, %
echolocation %
can detect distances to surfaces based on the reflections~\cite{sohl2015device,eliakim2018fully,villalpando2019ego,christensen2020batvision}.
Multi-channel audio is also used to track dynamic objects
\cite{murino,vehicle}.
Unlike any of these methods, our approach takes a video (both the audio and visual streams) as input and produces a floorplan as output.  Furthermore, our model is not restricted to known microphone layouts or known emitted sounds; rather, it can learn %
from natural sounds  sensed passively in the environment (e.g., running water, door shutting).
While environment semantics are explored in acoustic scene analysis~\cite{audioset,aasp}, our problem is quite different: the target output is a geometric map, not a label for the acoustic event that occurred.

\vspace*{-0.15in}
\paragraph{Audio-visual spatial sensing}

Audio and vision together offer powerful %
cues for spatial perception.  At the object level, %
they reveal shape and material properties~\cite{zhang2017generative,owens2016visually}, e.g., via the sound of one object striking another. At the environment level, audio can help sense 3D surfaces in cases where depth sensing would fail, e.g., transparent, shiny, or textureless surfaces~\cite{ye20153d,kim20173d}, or provide self-supervisory cues %
for imagery~\cite{visual-echoes}. 
Recent work leverages audio-visual sensing %
to address navigation tasks, learning to move efficiently to a sounding target~\cite{chen2019audio,gan2020look,chen2020waypoints}. In \cite{dean2020see}, a curiosity-driven framework is proposed that leverages audio-visual cues to explore environments.
None of the above methods produce audio-visual floorplans.  Furthermore, an insight unique to our work is the use of naturally occurring semantic sounds to understand a multi-room layout.

\begin{figure*}[h!]
    \centering
    \includegraphics[width=\textwidth]{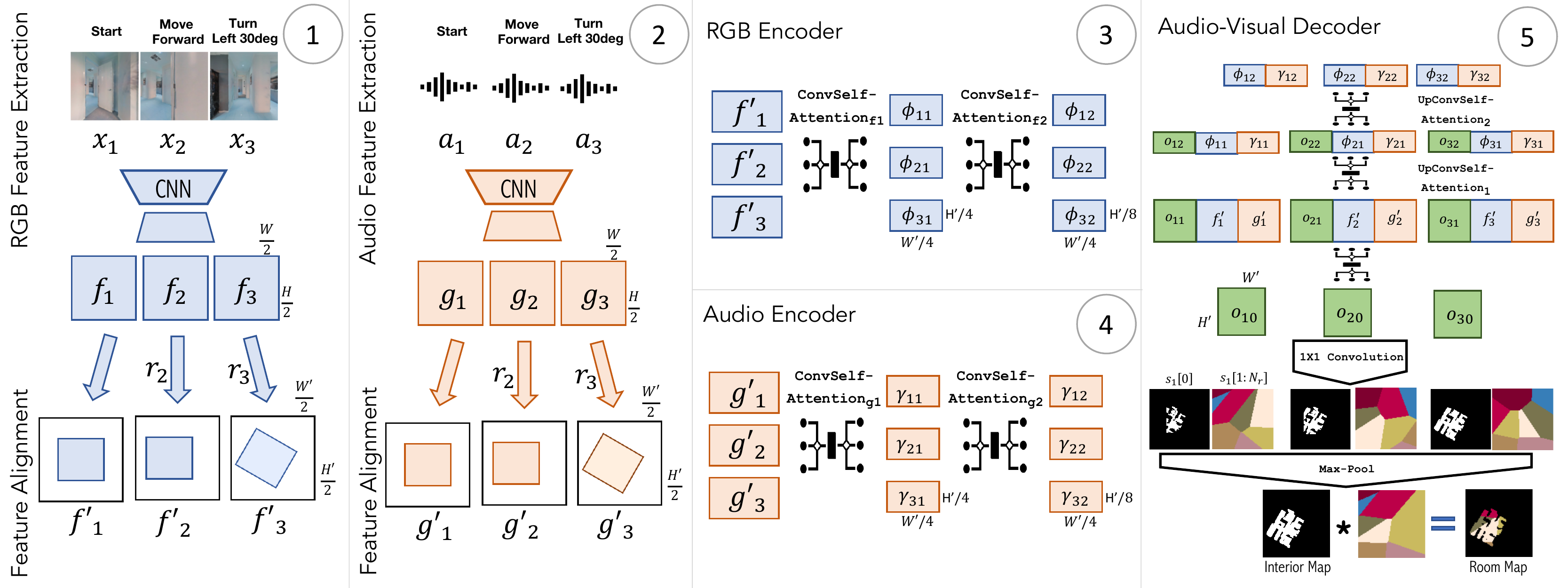}
    \vspace*{-0.25in}
    \caption{\textbf{\modelname model overview}: Our  model has three stages: feature extraction, feature alignment, and a sequence encoder-decoder. At each time step, the feature extractor obtains independent top-down floorplan-aligned features for each modality (ambisonic audio and RGB). These features are aligned to a common coordinate frame %
    using the relative motion of the camera. Entire sequences of audio and visual features are then processed by an encoder using a series of pixelwise self-attention operations and convolution layers. Finally, the two modalities are fused using a decoder architecture also including a series of self-attention and convolution layers. We depict training for three time steps here.  No matter how many steps are used in training, the model is applicable to arbitrary length sequences at test time. %
    \vspace*{-0.12in}
    }
    \label{fig:model}
\end{figure*}
\section{Approach}
\label{sec:approach}

Our goal is to estimate the 2D layout of an environment depicted in a short video. The 2D layout %
has two components: the structure of the interior area, and the semantic labels (room types) associated with each region of the interior. First, we formally describe the problem (Sec~\ref{subsec:setup}). We then describe our proposed model \modelname (Sec \ref{subsec:model}) %
and introduce our training and inference procedure (Sec \ref{subsec:training}).

\subsection{Problem Formulation}
\label{subsec:setup}
We consider videos generated by a camera and an ambisonic microphone following short trajectories through various home environments. An ambisonic mic captures omni-directional multi-channel audio~\cite{ward,rafaely}.
We represent a video by $V = \{(v_1, a_1), (v_2, a_2) ... \}$ where $v_t$ is the RGB frame and $a_t$ is the audio clip sampled at time step $t$. Additionally, let us denote by $P_V = \{\textbf{0}, {r_2}, {r_3}, ... \}$ the position of the camera and microphone relative to the first time step in the coordinate system of the floorplan, where $r_i = (x_i, y_i, \theta_i)$ represents the movement along the x- and y-axis on the 2D ground plane and $\theta_i$ represents the rotation about the gravity axis. Relative pose changes in a video can be estimated using computer vision~\cite{multiview}; for simplicity we assume correct relative camera poses are available for all methods. However, we find our proposed model is robust to noise in 
pose estimation 
within the range of odometry noise models considered in the literature~\cite{ramakrishnan2020occupancy}, owing to the resolution of the output floorplan.

Each floorplan is parameterized as two variables: $M_\text{int}$ and $M_\text{room}$, which represent the %
structure and semantics, respectively. The interior map $M_\text{int}$ is a 2D binary grid that is a top-down view of the environment and represents the existence of floor, objects, furniture by label $1$, and walls and areas outside the environment by label $0$. The room map $M_\text{room}$ is a 2D grid taking $N_r + 1$ possible values with labels $\{1,\ldots,N_r\}$ representing the $N_r$ room types (kitchen, bathroom, etc.) and $0$ representing walls and areas outside the environment. %
Each cell in the floorplan (an entry in the matrix $M$) represents a 25cm$^2$ area.

The goal of our work is to learn a mapping $\mathcal{F}$ that estimates the floorplan (both $M_\text{int}$ and $M_\text{room}$) of an environment using the video $V$ and the relative pose changes $P_{V}$. The visual information in $v_t$ captures the geometric properties and room types of visible regions. The audio information captured in $a_t$ is either actively emitted by the camera, or else passively generated by objects and people in the environment (details below). Since the placement of objects is highly correlated with room types (for example, showers are in bathrooms and dishwashers are in kitchens), the audio signal captures  a strong semantic signal indicating the room types. Furthermore, the echoes %
propagating through the environment capture geometric properties of the walls and other major surfaces. 
Our key insight is that the audio observations will illuminate the map for regions beyond what is visible in the frames of a short video.

\subsection{\modelname Floorplan Estimation Model}
\label{subsec:model}

We now present our \modelname floorplan estimation model $\mathcal{F}$. Fig.~\ref{fig:model} overviews our proposed model, consisting of three components: Top-Down Feature Extraction,
Feature Alignment,
and a Sequence Encoder-Decoder architecture. 
At each time step, 
\modelname estimates the interior map ($M_\text{int}$) and semantic room labels ($M_\text{room}$) in a %
neighborhood centered around the camera\KG{, integrating estimates over time.}

\vspace*{-0.14in}
\paragraph{Top-Down Feature Extraction}
The first stage of our proposed model involves extracting features for a given video $V$.  The purpose of the feature extraction step is to project egocentric visual frames $v_t$ and ambisonic audio clips $a_t$ to a 2D feature grid that is spatially aligned with the top-down floorplan that we wish to estimate at each time step. 

\emph{RGB Feature Extraction} ~ The feature extractor for an RGB frame $v_t$ consists of a ResNet-18 model up to \texttt{layer2} followed by a spatial pooling operation, which leads to a single 128-D feature. We use \texttt{layer2} features of the ResNet to capture low-level features like corners of rooms. This feature is then upsampled by a sequence of transposed convolutions leading to the final visual feature $f_t \in \mathbb{R}^{128\times \frac{H}{2}\times \frac{W}{2}}$ where $H,W$ are the height and width of the considered output floorplan area at each time step.  See Fig.~\ref{fig:model}, Column 1.
Importantly, this predicted area extends beyond the freespace directly observable from the visual frame $v_t$; in our experiments, the visually observed space on average covers only 13.7\% of the area in $H \times W$ when $H,W$ comprises of 40m$^2$ around the camera.
Below we explain how maps from multiple steps are aggregated for the video's final (larger) output map.

\emph{Audio Feature Extraction} ~ A video $V$  also consists of audio clips $a_t\in \mathbb{R}^{T\times 9}$ where $T$ is the sound duration at each time step $t$, and $9$ is the number of ambisonic channels \KG{corresponding to second order ambisonics}. Features for each audio clip are extracted using a sequence of linear, ReLU, and pooling layers, yielding a 128-D feature. Similar to the RGB extractor, this feature is upsampled via transposed convolutions to obtain the final audio feature $g_t \in \mathbb{R}^{128\times \frac{H}{2}\times \frac{W}{2}}$.  See Fig.~\ref{fig:model}, Column 2. 

\vspace*{-0.14in}
\paragraph{Feature Alignment}
After extracting features for the RGB frames and audio clips, each video is represented by the set of visual features $f_t$ and audio features $g_t$, for $t=1,\ldots,t_V$, where $t_V$ is the number of frames in video $V$ and may vary across videos. Note that each of these features was computed independently and thus far represents a feature aligned with the top-down map in a canonical \emph{frame-centric} coordinate frame. In order to process entire sequences, we need to establish correspondences between the features at each time step. Therefore, we next
align all the features to a common coordinate system, relative to the
first frame. In order to retain relative pose information, we concatenate a 64-channel 2D positional encoding map~\cite{vaswani2017attention} to each of the features (see Supp.~for more details). The aligned visual and audio features $f'_t,g'_t \in \mathbb{R}^{(128+64)\times \frac{H'}{2}\times \frac{W'}{2}}$ are computed by padding with zeros, and translating and rotating each feature $f_t,g_t$ by $r_t$ where $H'>H,W'>W$ due to padding.  See Fig.~\ref{fig:model}, bottom of Columns 1 and 2.

\vspace*{-0.1in}
\paragraph{Sequence Encoder-Decoder}
We now wish to encode features for each time step that account for information present in the entire sequence. For example, the appearance of a wall in the second frame should inform the features in the first step and vice-versa. Self-attention~\cite{vaswani2017attention} operations have shown to be useful to encode such bi-directional relationships. Inspired by this, we design a sequence of two self-attention and convolution operations (as shown in Fig.~\ref{fig:model}) which we refer to as $\mathtt{ConvSelfAttention}$. The self-attention operations are responsible for communication across time steps at each pixel \KG{location}. %
We use convolutions with stride 2 to also simultaneously downsample the features. We denote the resulting features for the RGB frames as:
\begin{gather}
\phi_{t1} = \mathtt{ConvSelfAttention_{f1}}(f'_t)\\    
\phi_{t2} = \mathtt{ConvSelfAttention_{f2}}(\phi_{t1}),
\end{gather}
where $\phi_{t1}, \phi_{t2}$ represent the intermediate encoder features. Similarly for the audio data, we represent by $\gamma_{t1}, \gamma_{t2}$ the outputs of the corresponding encoding layers $\mathtt{ConvSelfAttention_{g1}}$ and $\mathtt{ConvSelfAttention_{g2}}$. Note that since the convolutions downsample the features, we have $\phi_{t1}, \gamma_{t1}\in \mathbb{R}^{C_1\times \frac{H'}{4}\times \frac{W'}{4}}$ and $\phi_{t2}, \gamma_{t2} \in \mathbb{R}^{C_2\times \frac{H'}{8}\times \frac{W'}{8}}$.

So far, we have processed the visual and audio information independently. In order to take full advantage of the presence of both modalities, we allow cross-modal information transfer. We accommodate this in the final decoding stage of by concatenating the corresponding intermediate visual and audio features. For the decoder, we follow an architecture similar to the encoder by replacing the convolutions with transposed convolutions to upsample the features. We refer to these layers as $\mathtt{UpConvSelfAttention}$. More concretely, the decoder consists of three $\mathtt{UpConvSelfAttention}$ layers which are used to compute the output as:
\vspace*{-0.05in}
\begin{gather}
o_{t2} = \mathtt{UpConvSelfAttention_{2}}([ \phi_{t2}, \gamma_{t2} ])\\    
o_{t1} = \mathtt{UpConvSelfAttention_{1}}([ o_{t2}, \phi_{t1}, \gamma_{t1} ])\\
o_{t0} = \mathtt{UpConvSelfAttention_{0}}([ o_{t1}, f'_t, g'_t ]).
\end{gather}

\noindent
The final output $o_{t0} \in \mathbb{R}^{C\times H'\times W'}$ is classified using a 1x1 convolution giving us the final prediction at each time step as $s_t\in \mathbb{R}^{(N_r+1) \times H'\times W'}$. The first channel (notated as $s_t[0,:,:]$) represents a binary score map for the existence of interior space, and the remaining channels $1,2,\ldots N_r$ (notated as $s_t[i,:,:]$)  represent score maps for the existence of the corresponding room type. 
Note that due to the alignment step presented above,  these output maps are aligned in the common coordinate frame of the first time step. Therefore, to produce a prediction $S$ for the whole sequence, we max-pool the predictions $s_t$. The self-attention in the earlier encoder-decoder already accounts for communication across time steps for these per-step estimates.
\modelname outputs the aggregated interior and room classification scores for a video sequence:
$$S = \mathcal{F}(v_1,a_1, v_2, a_2, \ldots v_{t_V}, a_{t_V}). $$ 

In practice, for training, we fix the length of sequences $t_V=4$, which balances memory constraints with learning to integrate over time. For illustration, Fig.~\ref{fig:model} depicts an instance of the model with $t_V=3$. 

In summary, the proposed \modelname floorplan estimation model processes audio-visual sequences at various levels. The feature extraction independently processes each time step. The top-down alignment brings the features to a common coordinate frame. The encoders process sequences of each modality independently \KG{while integrating information across time}, and finally the decoder fuses information from both visual and audio modalities.

\subsection{Training and Inference}
\label{subsec:training}

The output of \modelname  
is a 2D map with $N_r + 1$ channels. The model is trained to predict two floorplan maps: the interior structure and the pixel-wise room labels. 

\vspace*{-0.1in}
\paragraph{Predicting Interior Maps}
Prediction of interior maps is a pixel-wise binary classification problem where $0$s represent the walls or \KG{exterior}  points %
and $1$s represent the points inside the environment (floors, furniture, objects, etc.).    
From %
$s_t$, the pixel-wise binary classification probability is computed using the sigmoid function: $p_{t_\text{int}}[i,j] = \frac{1}{1 + \exp {~ \-s_t\big[0,i,j~ \big]}}$
for each pixel location $i,j$ in the 2D grid. 

\vspace*{-0.1in}
\paragraph{Predicting Room Floorplans}
Prediction of room floorplans is similar to  prediction of interior maps, but requires multi-class classification of each pixel into one of $N_r$ semantic room types. Therefore, the class-wise probabilities at each pixel are computed using the softmax function. Concretely, the classification probability for class $n\in \{1,2,3,...,N_r\}$ at pixel location $(i,j)$ is computed as: 
\resizebox{0.7\linewidth}{!}{
$p_{t_\text{room}}[n-1, i,j] = \frac{\exp{~s_t\big[n,i,j \big]~ }}{\sum_{k=1}^{N_r} \exp{ ~s_t\big[k,i,j \big]~ }}.$
}

\vspace*{-0.1in}
\paragraph{Training Objectives}
For each time step $t$, let the ground truth interior and room maps of the $H\times W$ area around the camera be represented by $y_{t_\text{int}} \in \{0,1\}^{H\times W}$ and $y_{t_\text{room}} \in \{0,1,\ldots N_r\}^{H\times W}$. Since our model's predictions are aligned with time step $t=1$, we similarly align the ground truth maps to obtain $\hat{y}_{t_\text{int}}\{0,1\}^{H'\times W'}$ and $\hat{y}_{t_\text{room}}\in \{0,1,\ldots N_r\}^{H'\times W'}$ by padding with zeros, translating and rotating by $r_t$ (where $H', W'$ are the increased dimensions due to padding). The interior and room map classification objectives for each time step $t$ for pixel location $(i,j)$ are then defined as:
\vspace*{-0.05in}
\begin{equation}
\small
\mathcal{L}_{int} = %
\frac{1}{z} \sum_{t=1}^{t_V} \sum_{i=1}^{H'} \sum_{j=1}^{W'} -\hat{y}_{t_\text{int}}[i,j] \log p_{t_\text{int}} [i,j]
\end{equation}
\begin{equation}
\mathcal{L}_{room} = %
\frac{1}{z}\sum_{t=1}^{t_V}\sum_{i=1}^{H'} \sum_{j=1}^{W'} \sum_{k=1}^{N_r} -\mathbb{I}\big[ \hat{y}_{t_\text{room}}[i,j]{=}k\big]  \log p_{t_\text{int}} [k, i,j],
\end{equation}
where $z = t_V H' W'$,
$\mathbb{I}$ is the indicator function, and $t_V$ is the number of time steps in the video $V$. %
We ignore the unused pixel locations $(i,j)$ in $\hat{y}_t$ that arise from padding during the alignment step. \modelname  is trained using the sum of the two objectives: $\mathcal{L} = \mathcal{L}_{int} + \mathcal{L}_{room}$. 

During inference, we estimate the interior and room maps for the whole sequence. 
As explained above, this is done by max-pooling the %
predictions $s_t$ to produce a sequence-level prediction $S$.  
Importantly, the %
self-attention layers in our proposed model ensure that entire sequences are used to reason about each time step. Furthermore, since self-attention layers can process sequences of arbitrary length, we can apply the trained model on videos of varying length.

In order to predict the binary interior map, we simply threshold at $p=0.5$ the final pixel-wise interior probabilities. 
To obtain the room map prediction, %
\KG{we} assign the most likely room label to each location and use the thresholded interior map prediction  
as a binary mask \KG{to get its shape}. 

\subsection{Video Sequence Generation}
\label{subsec:data}

In order to generate videos in a variety of 3D environments for which we know ground truth floorplans, we use the Matterport3D dataset~\cite{chang2017matterport3d}\footnote{The Matterport3D license is available at http://kaldir.vc.in.tum.de/matterport/MP\_TOS.pdf.} %
and the SoundSpaces~\cite{chen2020soundspaces} audio simulations. SoundSpaces provides \KG{highly realistic} audio for 85 fully scanned real environments  
split 59/11/15 for train/val/test, respectively. Most environments are large multi-room homes and contain a variety of furnishings.  
SoundScapes provides precomputed impulse responses (IR) for all source-receiver locations on a dense grid sampled at 1m spatial resolution. \KG{The simulations use SoTA multi-band ray tracing, computing the IRs from arbitrary geometries and frequency-dependent acoustic material properties, and modeling both transmission (including through walls) and scattering.}  The IRs can be convolved with any audio clip to generate realistic audio for any chosen source-receiver location, including multiple simultaneous sources.  \KG{See~\cite{chen2020soundspaces} for details of the simulations and Supp.~videos for examples.}\\

\vspace*{-0.1in}
\noindent
\textbf{Generating Floorplans} ~ We use the Habitat-API~\cite{savva2019habitat} to generate top-down interior floorplans for each environment by projecting the point cloud %
to the 2D ground plane. %
Room floorplans are constructed using the Matterport3D room annotations by assigning a room label to each pixel of the interior floorplan. We use the  $N_r$=13 most frequent room labels from Matterport3D (laundry, kitchen, bathroom, etc.).\\

\vspace*{-0.1in}
\noindent
\textbf{Camera Trajectories} We generate videos %
by recording egocentric frames and ambisonic audio along short camera trajectories. Due to the grid constraint of the SoundSpaces data, we restrict camera positions to the same 1m grid. %
At each location, the camera is parallel to the ground plane and can have a rotation around the gravity-axis in the set $\{0^\circ, 30^\circ, 60^\circ, ... , 330^\circ \}$. 
At each step, while the RGB frame is constant, we record audio for 3 seconds. 

During training, the trajectories are randomly sampled. We train the models with fixed trajectory lengths of $t_V=4$ steps due to GPU memory constraints.  
In Supp.~we provide an ablation with $t_V=1$ to demonstrate the power of learning across the sequence.
For evaluation on the validation and test environments, we sample trajectories of variable length %
\KG{$t_V$, for $t_V \in \{1,2,4,8,16\}$.} \\

\vspace*{-0.1in}
\noindent
\textbf{Audio} We consider two settings of audio: \textit{device-generated} (active) and \textit{environment-generated} (passive). 
For the device-generated (Dev.~Gen.) setting, the video recording device (e.g., cell phone, AR headset, or robot) also emits a fixed recurring sound at each time step. %
We use a 3 sec frequency sweep chirp signal in the audible range (20Hz-20KHz).
Though any emitted sound could provide useful echoes, the wide range of frequencies activated in the sweep is expected to provide a particularly rich learning signal~\cite{jeong}.

In the environment-generated setting, rather than emit a sound, the system listens for %
naturally occurring sounds in homes. %
To achieve this, we first collect a set of 56/32/32 train/val/test audio clips\footnote{Downloaded from freesound.org} of duration 3 sec that capture sounds made by objects in different room types (for example, sound of a flush, dishwasher, etc.). This allows us to place  source sounds in the Matterport3D environments in the appropriate rooms. For each trajectory, the location of sound source(s) is randomly chosen, and the waveform played is dependent on the room type of that location.

We consider three \KG{passive} %
settings. %
In the first setting (referred to as Env.~Telephone), the source is near (within 40m$^2$ area) one of the steps in the trajectory and plays the ``telephone ring'' sound. In the second (Env.~Nearby) %
there is again a single sound source near the trajectory, but the audio clip varies according to the room type of the sampled source location. In the third (Env.~All Room), %
a source is randomly placed in each room and all sources simultaneously play a sound associated with their room type.

\section{Results}
\label{sec:results}

\begin{figure*}[t]
    \centering
    \includegraphics[width=\textwidth]{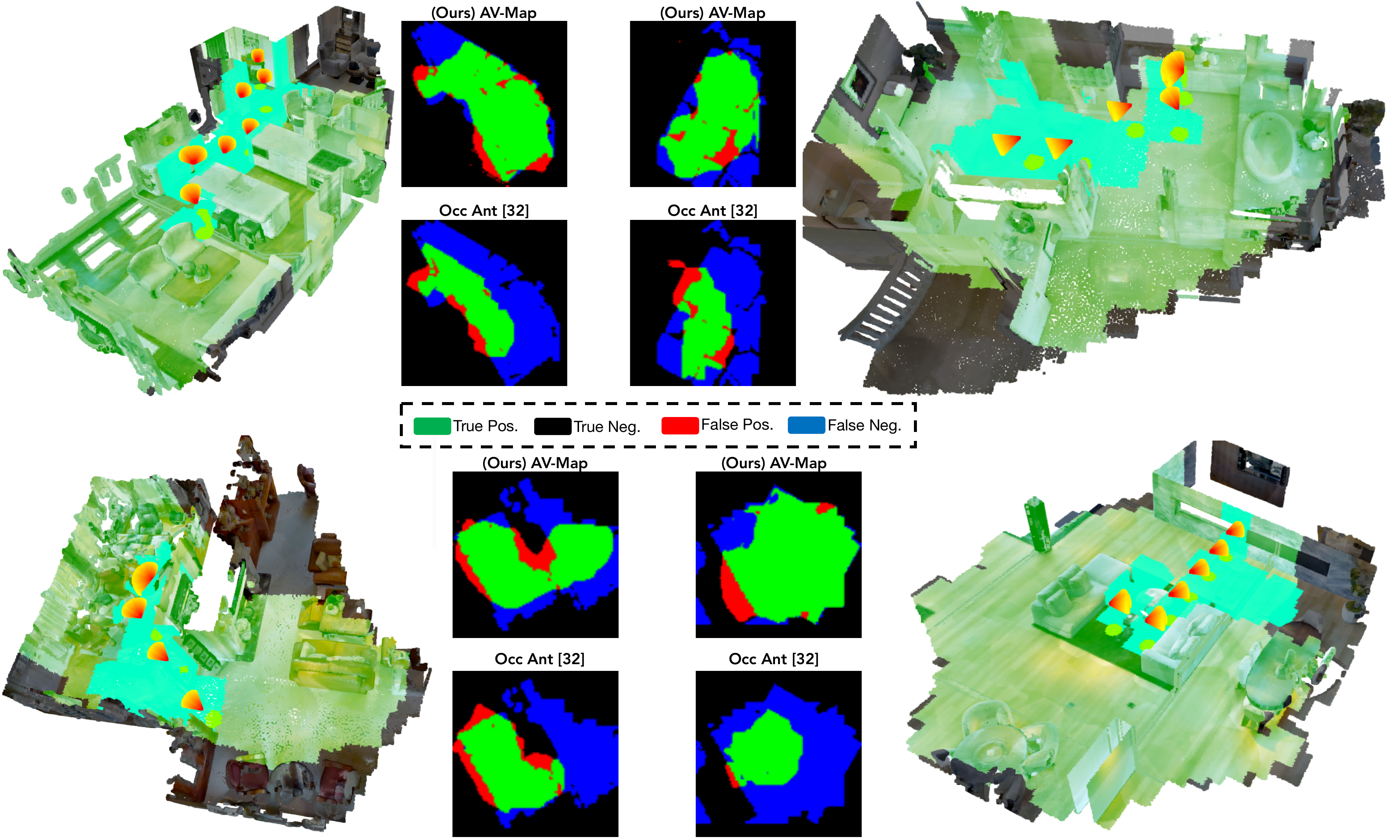}
    \caption{\textbf{Interior map predictions}: Here we visualize reconstructed interior maps estimated by OccAnt~\cite{ramakrishnan2020occupancy} and our proposed audio-visual model in the device-generated audio setting.  In the 3D examples, red cones are camera positions, cyan is the observed ground plane area, and transparent light green is the freespace inferred by our model.  2D output maps are color-coded against ground truth (see legend).  The perfect 2D map would be all green and black.  Our method ``sees" substantially more area by using audio, and it produces more accurate maps than the state-of-the-art mapping method~\cite{ramakrishnan2020occupancy} that also attempts to extrapolate beyond the directly observed area.}
    \vspace*{-0.15in}
    \label{fig:freespace_predictions}
\end{figure*}

\begin{figure}[h]
    \centering
    \includegraphics[width=\linewidth]{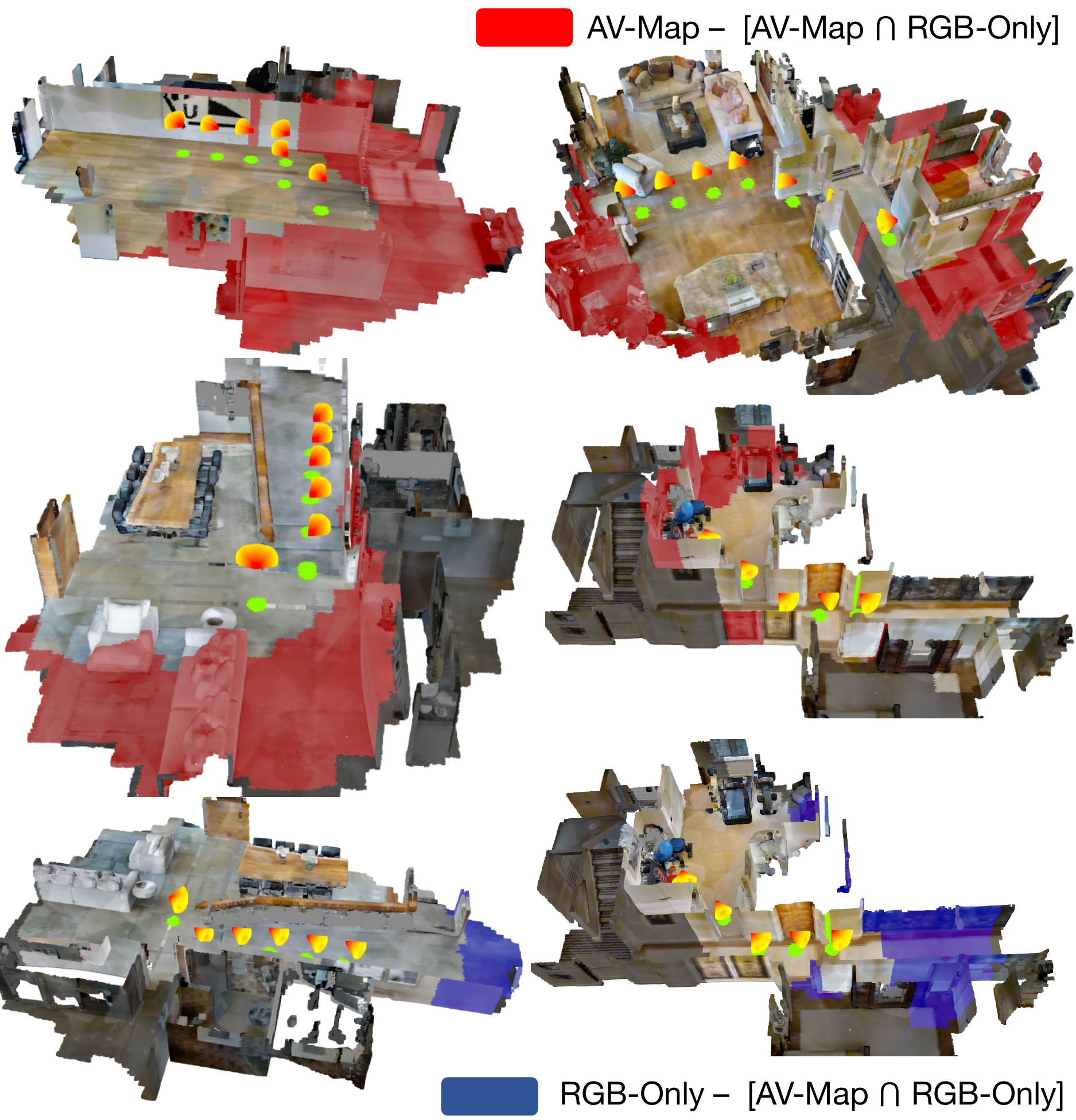}
    \caption{\textbf{\modelname vs.~RGB-only}: Here we compare the interior map predictions of \modelname and its RGB-only variant. In red, we show interior locations \KG{correctly} captured by \modelname \KG{but} not captured by the RGB-only model. In blue, we show locations captured by the RGB-only model \KG{but} missed by AV-Map. Observe that \modelname anticipates regions that are not visible through the camera (cone), which the RGB-only model cannot capture.
    }
    \vspace*{-0.1in}
    \label{fig:2d-interior}
\end{figure}

Through extensive qualitative and quantitative results, 
we demonstrate that our proposed model can effectively leverage both audio and visual signals to reason about the extent of the interior of environments (Sec \ref{subsec:freespace}) and classify regions of the interior into the associated rooms (Sec \ref{subsec:semantic}).

\vspace*{-0.15in}
\paragraph{Baselines}
In order to conduct a thorough analysis of our proposed model, we consider several baselines.

\noindent
\textbf{Interior-only:} simple baseline that predicts interior pixels (1s) everywhere in the considered neighborhood.

\noindent 
\textbf{Projected Depth:} standard occupancy map computed by projecting depth maps to the ground plane~\cite{chen2019learning,chaplot2020learning}.
Note that our model does not leverage depth, only RGB and audio.

\noindent
\textbf{OccAnt~\cite{ramakrishnan2020occupancy}:} The SoTA Occupancy Anticipation model \cite{ramakrishnan2020occupancy} 
infers a interior map (at each time step) for the 9m$^2$ area in front of the camera from RGB-D by learning to extrapolate beyond the visible ground-plane projections.  It is a key baseline to test our claim that audio can better ``see" beyond the visual observations.
We use the authors' code. 

\noindent
\textbf{Acoustic Echoes~\cite{dokmanic2013acoustic}:} This method assumes that all room shapes are convex polyhedra and estimates room shape by listening to audio echoes. However, this approach requires knowing the \emph{ground-truth} impulse responses at each microphone location, which our method does \emph{not} have access to.   %
While this method's setup is artificial, we 
use it as an upper bound for what an existing audio-only method could provide.

\noindent
\textbf{Ours audio-only and RGB-only:} 
As ablations of our model, we train variants with either modality removed.

Note that existing models like FloorNet~\cite{liu2018floornet,chen2019floor} are not applicable, since they require fully scanned point clouds as input.  In our setting, the input is simply a short sequence of  egocentric RGB views and audio.

\begin{table}[h]
\centering
\caption{\textbf{Interior reconstruction  evaluation:} Our proposed AV-Map model (here with device-generated sounds) outperforms existing methods and the baselines.  Methods creating only a binary map output cannot be scored by AP (NA).
}
\vspace*{-0.1in}
\label{tab:interior_quant_best}
\resizebox{0.85\linewidth}{!}{

\begin{tabular*}{\linewidth}{@{\extracolsep{\fill}}lccc@{}}
\toprule
 &                                            \textbf{AP}             & \textbf{Acc.}            & \textbf{Edge AP}        \\ \midrule
Interior-only                               & NA          & 50.00          & NA          \\
Projected Depth                             & NA             & 53.73          & NA             \\
OccAnt~\cite{ramakrishnan2020occupancy}     & 60.27          & 58.45          & 51.52          \\
Acoustic Echoes~\cite{dokmanic2013acoustic} & NA             & 50.37         & NA             \\ \midrule
(Ours) RGB-only                             & 69.05          & 64.20          & 52.93          \\
(Ours) Audio-only                           & 70.44          & 63.06          & 52.85          \\
(Ours) Audio-Visual                         & \textbf{73.67} & \textbf{66.51} & \textbf{55.21} \\ \bottomrule
\end{tabular*}

}\vspace*{-0.1in}
\end{table}

\subsection{Floorplan Interior Reconstruction}
\label{subsec:freespace}
First we present interior floorplan results.
We set $H\times W$ such that it covers 40m$^2$ area at each time step (see Supp.~for similar results with 164m$^2$).
 Since we aggregate the predictions from all time steps, the final accumulated area varies with the number of steps and direction of movement, adding at most 6.25m$^2$ at each step, for final output areas ranging from 40m$^2$ (1 step) to 134m$^2$ (16 steps). \\

\vspace*{-0.1in}
\noindent
\textbf{Evaluation Metrics} ~ 
We use three metrics: Average Precision (AP), Accuracy (Acc.), and Edge Average Precision (Edge AP). AP and Acc compare $S[0,:,:]$ and the binary ground truth map. Edge AP compares the edges of the predicted and ground truth maps in order to emphasize differences in boundary shapes. %
Pixels are reweighted in all metrics to balance the contribution of labels $0$ and $1$.\\

\vspace*{-0.1in}
\noindent
\textbf{Comparison to Baselines} Table~\ref{tab:interior_quant_best} presents our central result, a quantitative evaluation of all the baseline models on test trajectories of length $4$ steps in \KG{unseen} environments.
Our proposed AV-Map model using the device-generated audio outperforms all the baselines on all three metrics. Furthermore, our full AV model outperforms the audio-only and RGB-only variants by good margins. This shows our model successfully performs joint inference by leveraging important cues from both modalities. Our model with RGB-only is itself stronger than the baselines from existing literature~\cite{ramakrishnan2020occupancy,dokmanic2013acoustic}, showing the strength of our proposed framework even without the advantage of audio.  Fig.~\ref{fig:freespace_predictions} shows example map predictions compared to~\cite{ramakrishnan2020occupancy}, the best existing method.  They highlight how audio allows ``seeing" both behind the camera as well as inferring freespace behind walls in large multi-room homes.
Fig.~\ref{fig:2d-interior} compares examples from \modelname and its RGB-only variant.\\

\begin{figure}
    \centering
    \includegraphics[width=\linewidth]{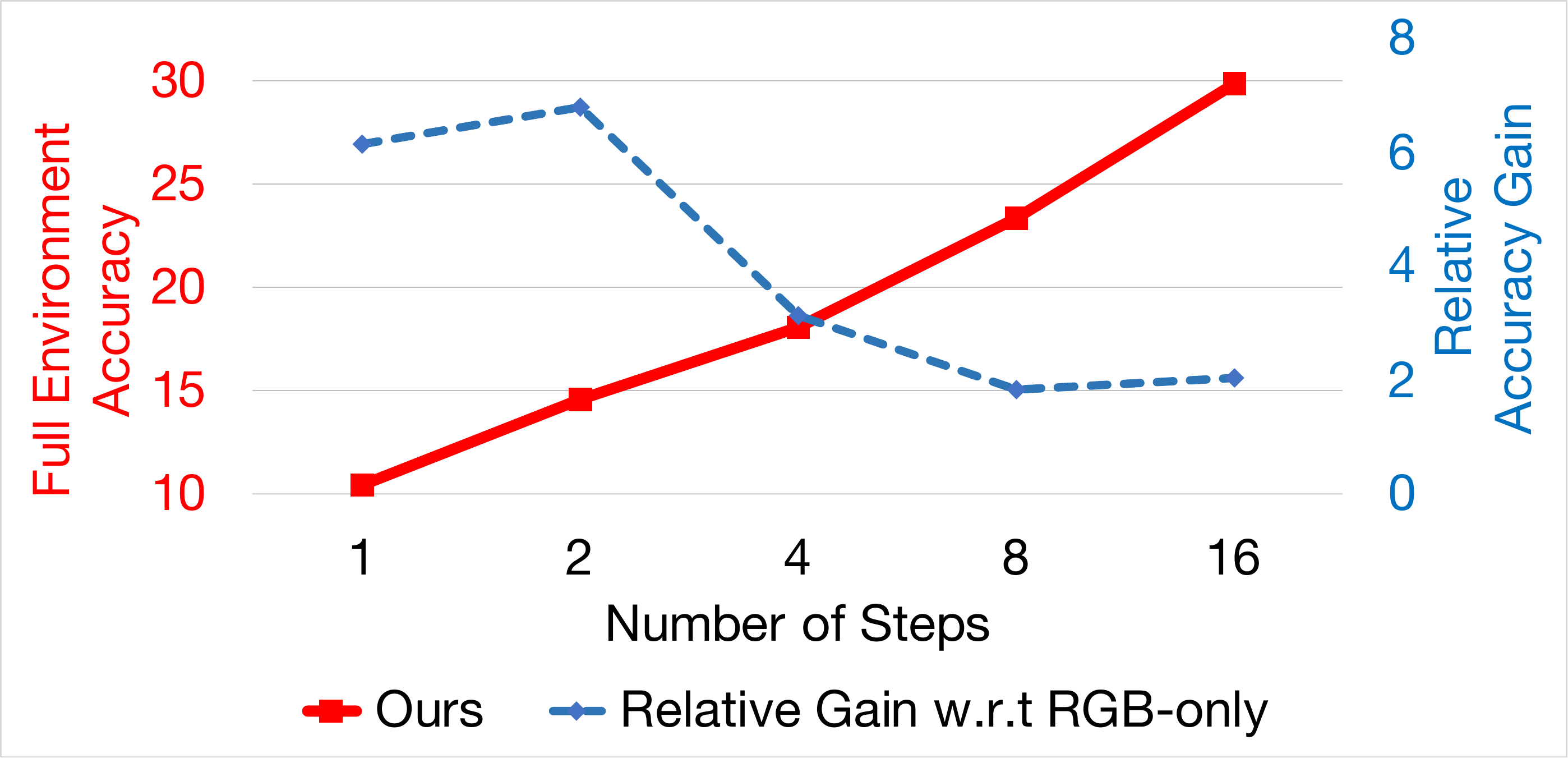}
    \caption{\textbf{Effect of trajectory length} %
   }\vspace*{-0.2in}
    \label{fig:iou_v_steps}
\end{figure}

\begin{table}[t]
\centering
\caption{\textbf{Interior reconstruction in different settings of audio:} Our proposed AV-Map model 
is applicable with either device-generated or environment-generated sounds.  
}
\label{tab:interior_quant_ablation}
\resizebox{0.9\linewidth}{!}{

\begin{tabular}{@{}llccc@{}}
\toprule
                                & & \textbf{AP}          & \textbf{Acc.}         & \textbf{Edge AP}       \\ \midrule
                                & RGB-only                         & 69.05       & 64.20       & 52.93         \\ \midrule
\multirow{2}{*}{Dev.~Gen.}      & Audio-only                       & 70.44       & 63.06       & 52.85         \\
                                & Audio-Visual                     & 73.67       & 66.51       & 55.21         \\ \midrule
\multirow{2}{*}{Env.~Telephone} & Audio-only                       & 68.60       & 63.53       & 53.27         \\
                                & Audio-Visual                     & 72.30       & 66.69       & 54.16         \\ \midrule
\multirow{2}{*}{Env.~Nearby}    & Audio-only                       & 66.27       & 61.49       & 51.86         \\
                                & Audio-Visual                     & 72.86       & 66.16       & 54.32         \\ \midrule
\multirow{2}{*}{Env.~All Room}  & Audio-only                       & 65.42       & 61.41       & 51.98         \\
                                & Audio-Visual                     & 71.38       & 65.46       & 54.79         \\ \bottomrule
\end{tabular}
}
\vspace*{-0.1in}
\end{table}

\vspace*{-0.15in}
\noindent
\textbf{Comparison of Audio Settings:} Table~\ref{tab:interior_quant_ablation} compares %
our \modelname model in the three audio settings described in Sec~\ref{subsec:data}. %
Our model performs slightly better when operated in the device-generated audio setting. This is expected since the frequency sweep audio allows us to capture all the frequencies in the audible range, unlike the naturally occurring sounds in the environment-generated settings. Furthermore, the relative location of the source is known in the device-generated setting (since it is always at the camera). 
In every setting, \modelname outperforms the RGB-only and Audio-only ablations. %
Despite the challenges in Env.~All Room (multiple simultaneous sounds from different source locations), we observe minimal decline in our interior map performance.\\

\noindent
\textbf{Effect of Trajectory Length:}
Figure~\ref{fig:iou_v_steps} shows our \emph{full-house} accuracy gains as a function of trajectory length.\footnote{Absolute numbers are lower than in Tab.~\ref{tab:interior_quant_best} because the scored area here is the entire house's area, vs.~the maximum output map area in Tab.~\ref{tab:interior_quant_best}.}
Our  model
outperforms the RGB-only %
ablation consistently across varying trajectory lengths. Importantly, our gains are largest when the video is shorter, when %
less area is visible.  This again confirms the power of audio to ``see" beyond the images. As the length increases, the visible fraction of the environment becomes larger, diminishing the impact of audio signals (dotted blue line). %

\begin{figure}[h!]
    \centering
    \begin{subfigure}[t]{\linewidth}
    \centering
    \includegraphics[width=\linewidth]{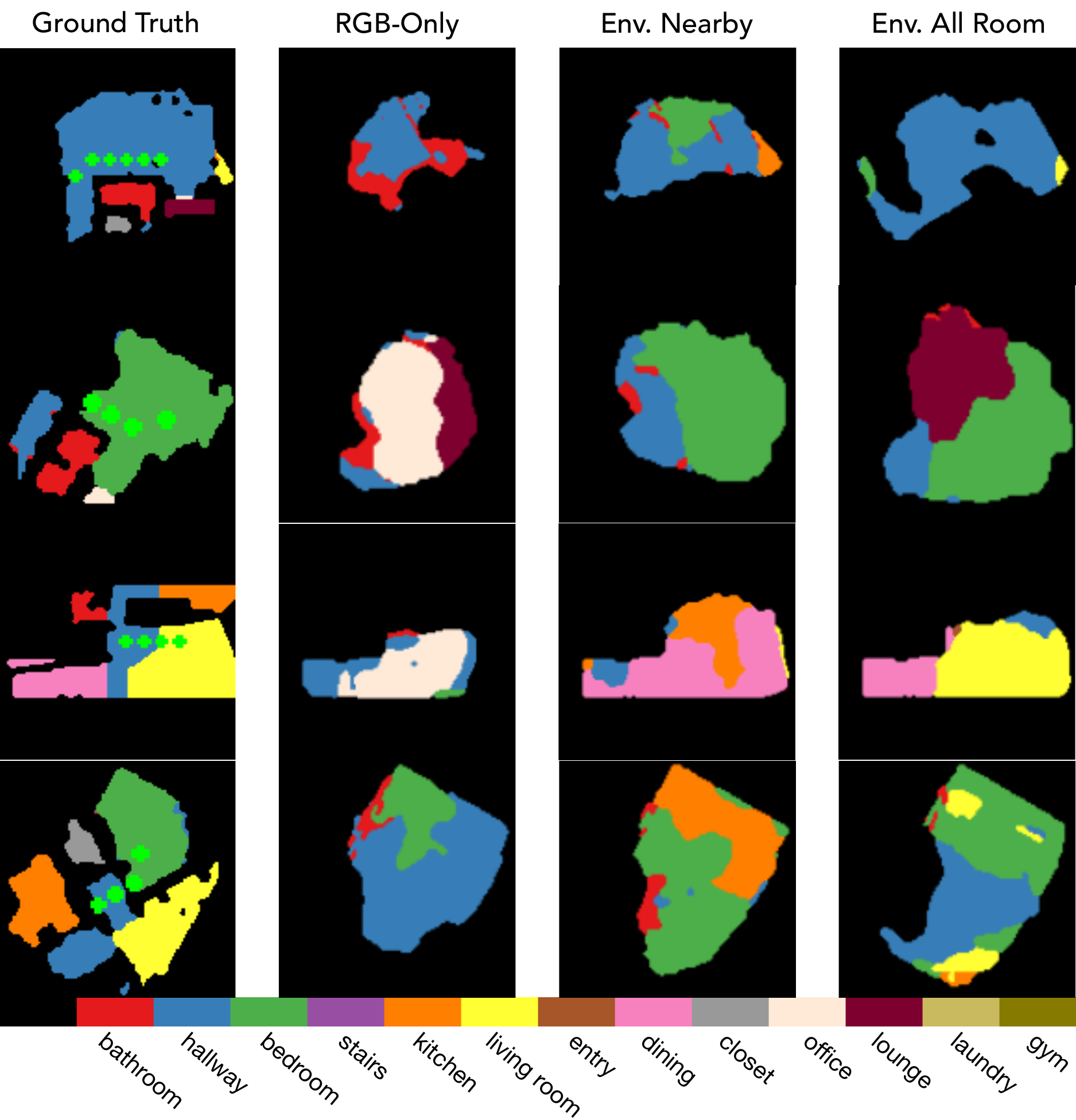}
    \end{subfigure}
    
    \begin{subfigure}[t]{\linewidth}
    \centering
    
\begin{tabular}{@{}lccc@{}}
\toprule
        & RGB-Only  & Audio-only  & AV-Map      \\ \midrule
Mean AP & 9.42 & 13.30 & \textbf{14.12} \\ 
\bottomrule
\end{tabular}

    \end{subfigure}
    \caption{\textbf{Room map predictions:} Our model infers the general layout of rooms in the home based on what it sees in the short video as well as the natural occurring sounds it hears.   Green dots on ground truth maps are camera positions.
    }
    \label{fig:room_predictions}
    \vspace*{-0.1in}
\end{figure}
\subsection{Floorplan Room Classification}
\label{subsec:semantic}

Finally, we evaluate the semantic room label maps. %
We evaluate with environment-generated audio, which provides natural object-room cues.  %
Figure~\ref{fig:room_predictions} compares the mean of the pixel-wise room classification average precision for our proposed model in the Env. Gen. All Room setting and its ablations.\footnote{Note that the baselines from Table~\ref{tab:interior_quant_best} are not applicable here because they produce only geometric interior maps.}
Results are averaged over all trajectory lengths.
Audio can identify rooms in the neighborhood of the trajectory, and we see best results when both modalities are used together. The room map examples show
\modelname provides better room classification compared to our RGB-only variant, and does best with many natural semantic sounds---encouraging for deployment in a busy household. %
Our model can identify the correct room type and its approximate location, though without actually entering a room, its exact footprint naturally remains ambiguous. %

\section{Conclusion and Future Work}
\label{sec:conclusion}

We proposed a new research direction: audio-visual floorplan reconstruction from short video sequences. We developed a multi-modal model %
to estimate the floorplan around and far beyond the camera trajectory.  Our AV-Map model successfully 
infers the structure \KG{and semantics} of areas that are not visible, outperforming the state-of-the-art in extrapolated visual maps. %
\KG{In future work we plan to consider extensions to multi-level floorplans and connect our mapping idea to a robotic agent actively controlling the camera.}

\section{Acknowledgements}
\label{sec:ack}

We would like to thank Unnat Jain, Changan Chen, and Santhosh Kumar Ramakrishnan for help with generating the audio simulations and providing help with code for this work. We would also like to thank Ruohan Gao for providing feedback on the text. 

{\small
\bibliographystyle{ieee_fullname}
\bibliography{papers}
}

\clearpage
\onecolumn

\begin{appendix}
\begin{center}
  \large\textbf{(Supplementary Material) Audio-Visual Floorplan Reconstruction }\\
\end{center}

\section{Additional Interior Map Visualizations}
\label{suppsec:interior_vis}

\begin{figure}[h!]
    \centering
    \includegraphics[height=0.8\textheight]{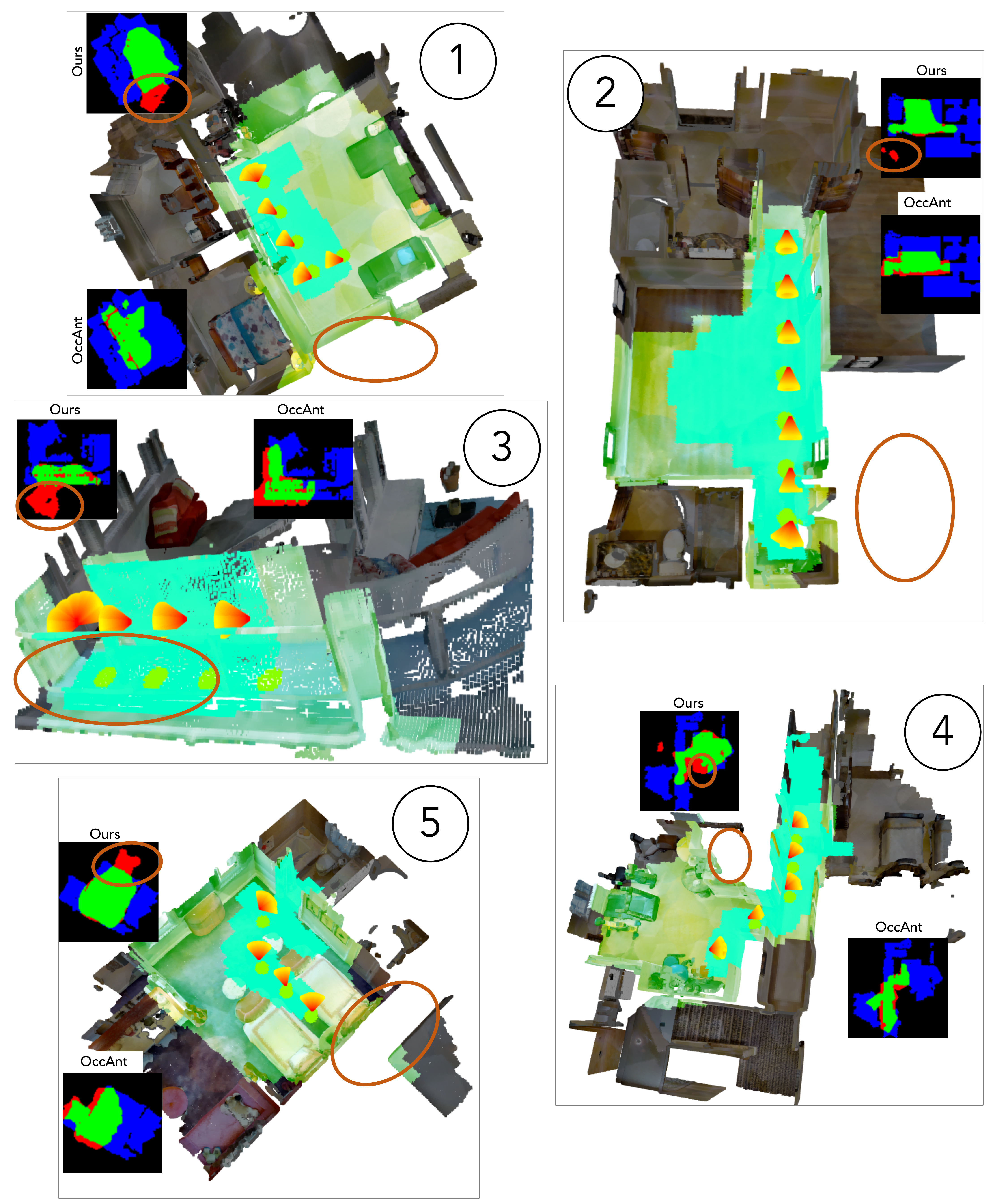}
    \caption{\textbf{Additional Interior Map Visualizations:}  We present additional visualizations of the estimated Interior Maps. The circled areas indicate the locations that are misclassified by our proposed model. See text for discussion.
   }\vspace*{-0.2in}
    \label{fig:extra_interior_vis}
\end{figure}

Figure~\ref{fig:extra_interior_vis} presents additional AV-Map interior map prediction visualizations, like Fig 3 in the main text.  We see again how our model sees beyond the visible portions (cyan) to more fully map the space.  We also highlight our failure modes; see the mis-classified locations (circled) on the predicted maps. We observe that the errors often arise in challenging locations that are not visually covered, where the model relies on the audio signal (see Figure~\ref{fig:extra_interior_vis} sample 1,4,5). Some errors arise from noise in the scan of the environment (see Figure~\ref{fig:extra_interior_vis} sample 3 - missing point cloud) since the rendered RGB frames are noisy. 

\section{Room Map Visualizations and Confusion Analysis}
\label{suppsec:semantic_vis}

\begin{figure}[h!]
    \centering
    \includegraphics[height=0.75\textheight]{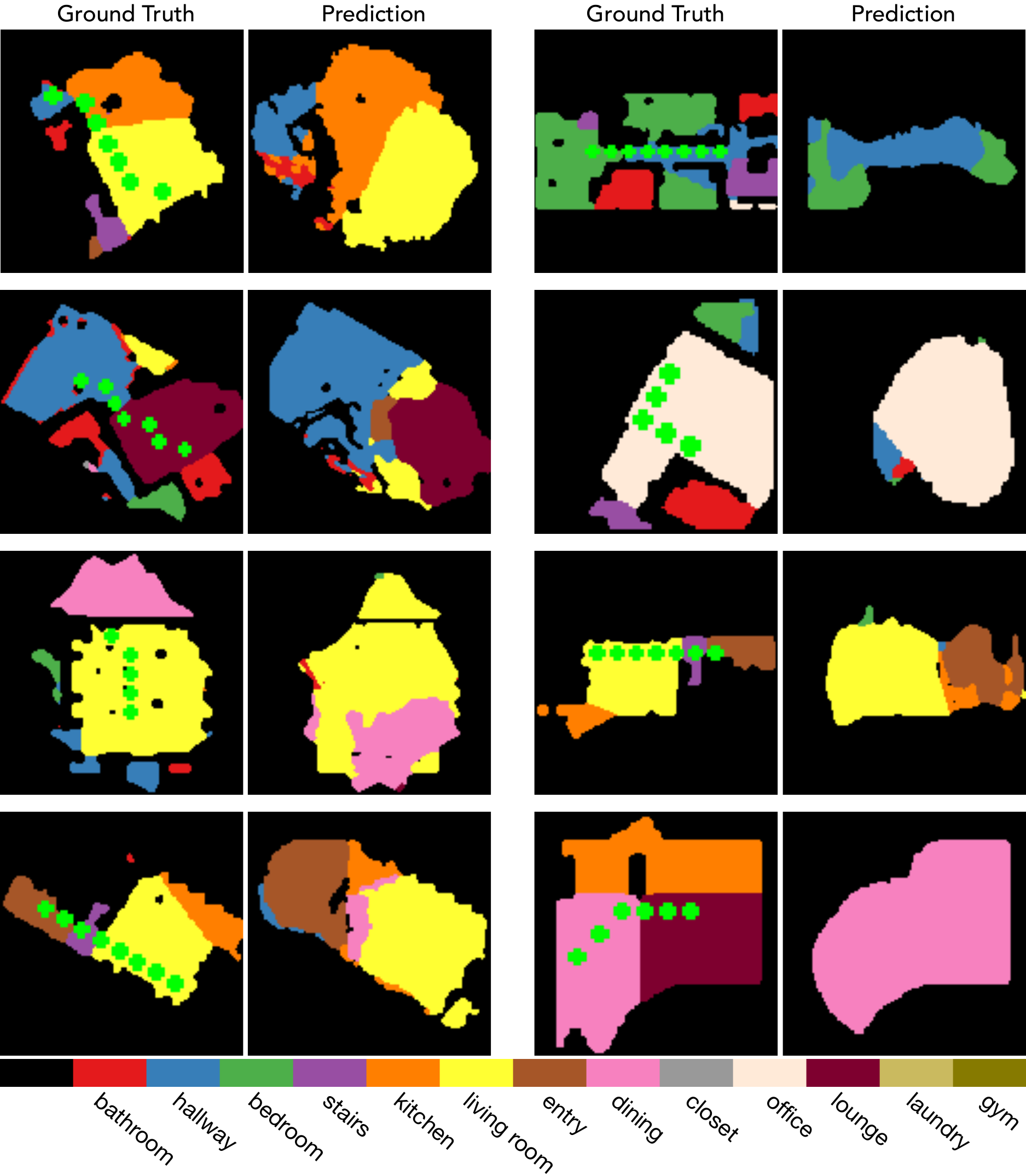}
    \caption{\textbf{Additional Room Map Visualizations} 
   }
    \label{fig:extra_semantic_vis}
\end{figure}

In Figure~\ref{fig:extra_semantic_vis}, we present additional visualizations for the estimated room maps. The room maps were generated by the \modelname model operating in the environment generated all-room audio setting.  Green dots on the ground truth indicate the camera positions. From these visualizations, we observe that the model can successfully identify the approximate locations of several rooms. Some sources of errors are errors in interior estimation (see Column 1, Row 4 and Column 2, Row 3) and errors in localization of the rooms (see Column 1, Row 3).

\begin{figure}[h!]
    \centering
    \includegraphics[width=0.85\textwidth]{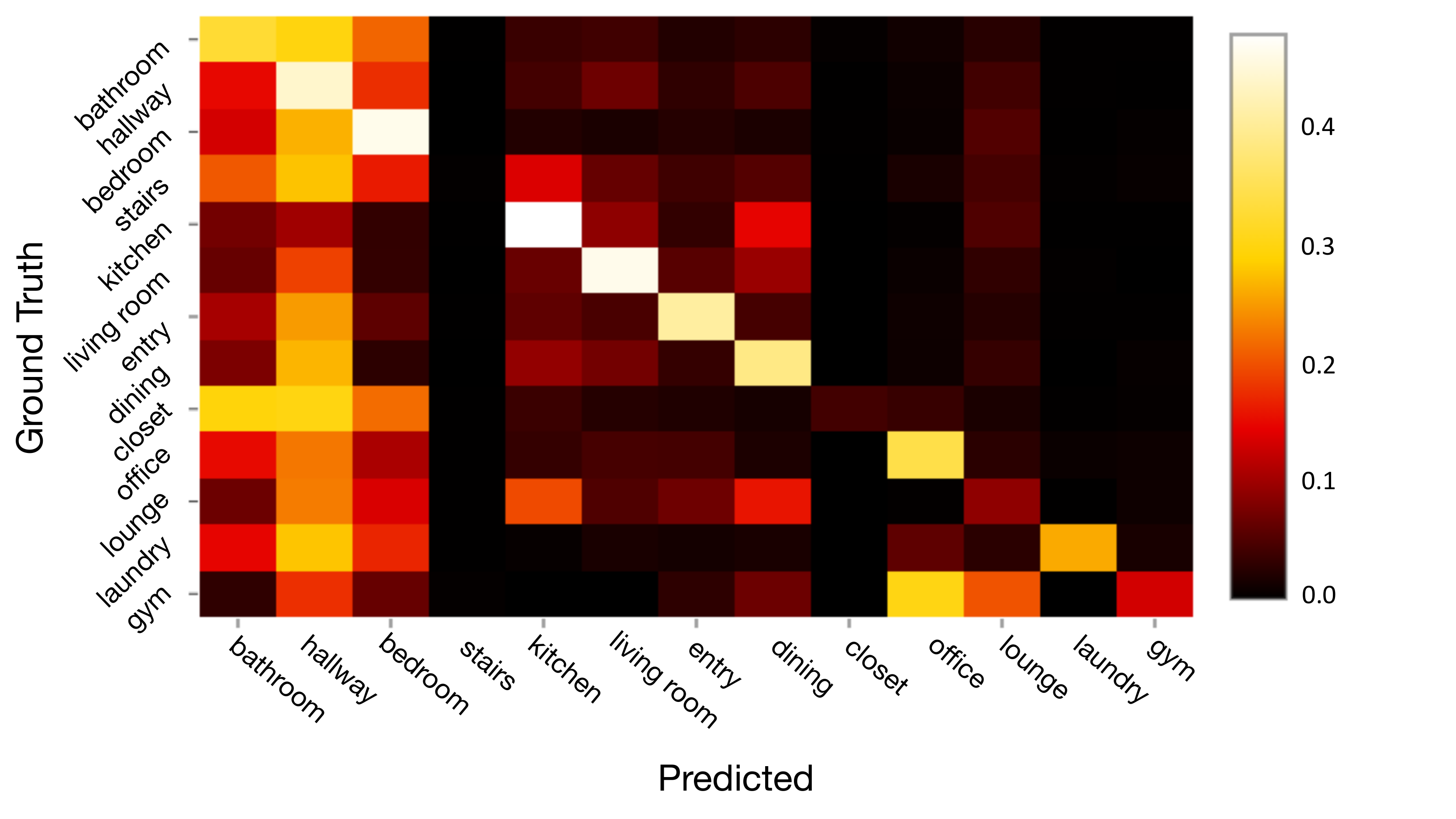}
    \caption{\textbf{Room Class Confusion Matrix} 
   }
    \label{fig:confusion}
\end{figure}

In Figure~\ref{fig:confusion}, we present a confusion matrix for the pixel-wise room label predictions. We observe that there is a bias towards predicting the ``bathroom'', ``hallway'' and ``bedroom'' classes which are the three most frequent room labels. The two least frequent classes (``stairs'' and ``closet'') are almost never predicted. This indicates that our model could benefit from training on a larger, more diverse and more balanced dataset. We also find that the rooms that are usually in close proximity have slightly higher confusion rates - for example, bedroom vs bathroom, and dining room vs kitchen. This suggests that our model struggles to accurately localize the boundaries of rooms (as also indicated in the main text).

\section{Importance of Sequence Modeling}
\label{suppsec:seq_modeling}

At each time step of a video, the audio clip $a_i$ is generated by convolving a downloaded audio clip $c$ with an impulse response $\omega_i$. Therefore, the audio clip can be expressed as:
\begin{gather}
    a_i = c \circledast \omega_i \\
    \text{OR} \nonumber\\
    \mathbb{F}(a_i) = \mathbb{F}(c) \mathbb{F}(\omega_i) 
\end{gather}

\noindent
where $\mathbb{F}$ is the Fourier transform. The impulse response encodes the acoustic characteristics of the environment for the given source location and receiver location pair at the time step $i$. These acoustic characteristics of the environment strongly depend on the geometric and material properties of the environments. Therefore, in order to infer the geometric properties of the environment, a model should ideally be able to either disentangle the impulse response $\omega_i$ from the audio clip $a_i$ or infer a function of the impulse response from $a_i$. This is not possible from the audio clip at a single time step unless the audio clip $c$ is known apriori. However, listening to audio clips from multiple time steps $a_i$, $a_j$, can allow us to model relative changes in impulse responses as:
\begin{gather}
    \frac{\mathbb{F}(a_i)}{\mathbb{F}(a_j)} = \frac{\mathbb{F}(\omega_i)}{\mathbb{F}(\omega_j)} = \text{relative change in impulse response}
\end{gather}
These relative changes can also provide information about the geometric properties of the environment. For example, walking past a door of a room containing a sound source will see a large change in impulse response clearly indicating the presence of an opening. Note here that the inferred relative change in impulse response does not rely on the original audio clip $c$ anymore. This is also a favorable feature since downloaded audio clips are not 100\% anechoic. So in practice the audio clips $c$ encode the acoustic characteristics of the recording environment \textit{i.e.} $c=\hat{c} \circledast \omega_{rec}$ where $\hat{c}$ is the anechoic audio and $\omega_{rec}$ is the impulse response of the recording setup. While we do not explicitly enforce the \modelname model to infer these relative changes, training with multiple audio clips forces the model to learn to disentangle the effect of the impulse response. 

The proposed \modelname model allows training and testing with video sequences of arbitrary length. During training, the primary bottleneck for using very long sequences is the memory footprint and speed of computation. Training with $t_V=1$ is equivalent to making independent predictions at each time step and pooling them to obtain the final interior map estimate. For each time step, such a model would not be able to make inferences using visual features in other time steps (for example, the fact that the camera entered a door in the first step provides additional context at the second time step). Furthermore, as explained above, making independent predictions does not allow us to model relative changes in the impulse responses. We observed that sequences of length $t_V=4$ provide the benefits of modeling sequences while maintaining a manageable training duration. As promised in the main paper, in Table~\ref{tab:step_v_seq}, we show results with $t_V=1$ and compare to the $t_V=4$ setting to demonstrate the impact of sequence modeling. 

\begin{table}[h!]
\centering
\caption{\textbf{Impact of Sequence Modeling:} We observe a significant improvement in performance of our \modelname model when trained on sequences compared to making independent predictions at each step.
}
\vspace*{-0.1in}
\label{tab:step_v_seq}

\begin{tabular}{@{}lccc@{}}
\toprule
 &  \textbf{AP}             & \textbf{Acc.}            & \textbf{Edge AP}        \\ \midrule
Single Step ($t_V=1$)                    &  68.91 & 62.46  & 54.05 \\
Sequence ($t_V=4$)                       &  \textbf{73.61} & \textbf{66.51}  & \textbf{55.21}  \\
\bottomrule
\end{tabular}

\end{table}
\section{Predicting $164m^2$ interior area}
\label{suppsec:area164}

In the main text, we presented a quantitative analysis of the \modelname model trained to estimate interior maps for an area of 40$m^2$ around the camera at each time step (by setting hyper-parameters $H,W$). As promised in Section 4.1 of the main paper, here in Table~\ref{tab:ap_area164}, we present similar quantitative results\footnote{Note that the positive and negative pixels are balanced by reweighting as discussed in Section 4.1} for a model trained to predict a 164$m^2$ area around the camera at each step. We observe similar results demonstrating the improved performance of the \modelname model compared to the RGB-only model.

\begin{table}[h]
\centering
\caption{\textbf{Interior Map Average Precision}: We present a qualitative analysis of various models trained to predict an interior area covering $164m^2$ at each time step.}
\label{tab:ap_area164}

\begin{tabular}{@{}lccccc@{}}

\toprule
Number of Steps $\xrightarrow[]{}$          & 1              & 2              & 4              & 8              & 16             \\ \midrule
RGB only      & 72.05          & 72.45          & 72.60          & 75.05          & 76.00          \\
Dev. Gen.     & \textbf{73.59} & \textbf{75.00} & \textbf{75.10} & \textbf{78.75} & \textbf{80.71} \\
Env. Nearby   & 73.01          & 73.36          & 74.08          & 78.03          & 79.76          \\
Env. All Room & 72.33          & 73.55          & 74.76          & 77.67          & 80.29          \\ \bottomrule
\end{tabular}
\end{table}

\section{Additional Dataset Details}
\label{suppsec:data_details}

\subsection*{Environments}
We use the Matterport3D\cite{chang2017matterport3d} dataset to generate video sequences (see Sec 3.4 of the main text). We use the splits provided by the SoundSpaces~\cite{chen2020soundspaces} dataset for training, validation, and testing. We include the environments in the splits here for reference: \\

\noindent
\textbf{Train environments}: ['17DRP5sb8fy', '1LXtFkjw3qL', '1pXnuDYAj8r', '29hnd4uzFmX', '5LpN3gDmAk7', '5q7pvUzZiYa', '759xd9YjKW5', '7y3sRwLe3Va', '82sE5b5pLXE', '8WUmhLawc2A', 'aayBHfsNo7d', 'ac26ZMwG7aT', 'B6ByNegPMKs', 'b8cTxDM8gDG', 'cV4RVeZvu5T', 'D7N2EKCX4Sj', 'e9zR4mvMWw7', 'EDJbREhghzL', 'GdvgFV5R1Z5', 'gTV8FGcVJC9', 'HxpKQynjfin', 'i5noydFURQK', 'JeFG25nYj2p', 'JF19kD82Mey', 'jh4fc5c5qoQ', 'kEZ7cmS4wCh', 'mJXqzFtmKg4', 'p5wJjkQkbXX', 'Pm6F8kyY3z2', 'pRbA3pwrgk9', 'PuKPg4mmafe', 'PX4nDJXEHrG', 'qoiz87JEwZ2', 'rPc6DW4iMge', 's8pcmisQ38h', 'S9hNv5qa7GM', 'sKLMLpTHeUy', 'SN83YJsR3w2', 'sT4fr6TAbpF', 'ULsKaCPVFJR', 'uNb9QFRL6hY', 'Uxmj2M2itWa', 'V2XKFyX4ASd', 'VFuaQ6m2Qom', 'VVfe2KiqLaN', 'Vvot9Ly1tCj', 'vyrNrziPKCB', 'VzqfbhrpDEA', 'XcA2TqTSSAj', 'D7G3Y4RVNrH', 'E9uDoFAP3SH', 'JmbYfDe2QKZ', 'r1Q1Z4BcV1o', 'r47D5H71a5s', 'ur6pFq6Qu1A', 'VLzqgDo317F', 'YmJkqBEsHnH', 'ZMojNkEp431']

\noindent
\textbf{Val environments}: ['2azQ1b91cZZ', '8194nk5LbLH', 'EU6Fwq7SyZv', 'oLBMNvg9in8', 'QUCTc6BB5sX', 'TbHJrupSAjP', 'X7HyMhZNoso', 'pLe4wQe7qrG', 'x8F5xyUWy9e', 'Z6MFQCViBuw', 'zsNo4HB9uLZ']

\noindent
\textbf{Test environments}: ['5ZKStnWn8Zo', 'ARNzJeq3xxb', 'fzynW3qQPVF', 'jtcxE69GiFV', 'pa4otMbVnkk', 'q9vSo1VnCiC', 'rqfALeAoiTq', 'UwV83HsGsw3', 'wc2JMjhGNzB', 'WYY7iVyf5p8', 'YFuZgdQ5vWj', 'yqstnuAEVhm', 'gxdoqLR6rwA', 'gYvKGZ5eRqb', 'Vt2qJdWjCF2']

\subsection*{Room types and associated sounds}
For generating room maps, we choose the 13 most frequent room types. For each room type, we download sounds from \url{www.freesound.org} generated by objects (or people) that are unique to the room type. Here we present the list of rooms, their associated sounds, and the number of train/val/test sounds for each:
\begin{itemize}
\itemsep 0em
    \item bathroom: brushing (3/2/2), flush(4/1/1)
    \item hallway: <no sound>
    \item bedroom: alarm clock (5/3/3)
    \item stairs: footsteps (5/3/3)
    \item kitchen: blender (3/1/1), cabinet (1/1/1), dishwasher (3/2/2)
    \item living room: telephone (5/3/3)
    \item entryway/foyer/lobby: knock (5/2/2)
    \item dining room: knife (4/1/1)  , spoon(4/2/2) 
    \item closet: closet door (2/2/2)
    \item office: keyboard (5/3/3)
    \item lounge: no sound
    \item laundryroom/mudroom: washing machine (5/3/3)
    \item workout/gym/exercise: person panting (5/3/3)
\end{itemize}

\section{Implementation and Training Details}
\label{suppsec:training_details}

\subsection{Hyperparameters}
The \modelname model is trained with a batchsize of 32 videos using 4 GPUs. Each sample in the batch is generated by randomly sampling a camera trajectory as described in the main text. We use the SGD optimizer with a starting learning rate of 0.1, momentum 0.9 and weight decay 0.00001. After 30000 SGD updates, we drop the learning rate to 0.01 and train for an additional 20000 SGD steps. 

\subsection{Positional Encoding}
The positional encoding map added in the feature alignment stage (see Sec 3.2) is a 64-channel 2D map representing the position of each pixel with a 64 dimensional vector. For position (i,j) in the feature map, the positional encoding $PE(i,j)$ is computed as:

\begin{multline*}
    PE(i) = \Big[ \sin(\frac{i}{10000^{0/32}}), \cos(\frac{i}{10000^{0/32}}), \sin(\frac{i}{10000^{2/32}}), \cos(\frac{i}{10000^{2/32}}), \ldots \\
    \sin(\frac{i}{10000^{30/32}}), \cos(\frac{i}{10000^{30/32}}) \Big]\\
\end{multline*}
\begin{align}
    PE(i,j) = [PE(i), PE(j)] 
\end{align}

\subsection{Feature Alignment}
Here we present a pseudo-code to illustrate the feature alignment described in Section 3.2. 
\begin{algorithm}[ht]
\label{alg:align}
\definecolor{codeblue}{rgb}{0.25,0.5,0.5}
\lstset{
  backgroundcolor=\color{white},
  basicstyle=\fontsize{7.2pt}{7.2pt}\ttfamily\selectfont,
  columns=fullflexible,
  breaklines=true,
  captionpos=b,
  commentstyle=\fontsize{7.2pt}{7.2pt}\color{codeblue},
  keywordstyle=\fontsize{7.2pt}{7.2pt},
  numbers=right,
  mathescape=true
}
\begin{lstlisting}[language=python]
# f_t: visual features at time-step t 
# g_t: audio features at time-step t 
# r_t=(x_t, y_t, $\theta$_t): relative position in meters and angle 
# res: per-pixel feature resolution (of f_t,g_t) in meters

max_feat_dim = max(f_t.shape[-1],f_t.shape[-2])
max_disp = max([x_1/res, x_2/res, ..., x_$t_v$/res, y_1/res, y_2/res, ..., y_$t_v$/res])

# Calculate amount to pad 
# = maximum displacement.+ sqrt(2)* feature dimension
# 2nd term accounts for rotation by 45$^\circ$
padding = max_disp +  max_feat_dim*sqrt(2) 


# def concat_position_embedding: append 64 position embedding channels (see Sec F.2)
# def pad(f,n): expand each spatial dimension by 2n (n before and n after) and fill with zeros
# def translate(f, (x,y)): move the feature vertically by y pixels and horizontally by x pixels
# def rotate(f, $\theta$): rotate feature by $\theta$ about the center of the feature map


for t in 1,2,...,t_V:
    f_t = concat_position_embedding(f_t)
    f$'$_t = pad(f_t,padding)
    f$'$_t = translate(f$'$_t, (x_t/res, y_t/res))
    f$'$_t = rotate(f$'$_t, $\theta$_t)
    

for t in 1,2,...,t_V:
    g_t = concat_position_embedding(g_t)
    g$'$_t = pad(g_t,padding)
    g$'$_t = translate(g$'$_t, (x_t/res, y_t/res))
    g$'$_t = rotate(g$'$_t, $\theta$_t)


\end{lstlisting}
\end{algorithm}

\end{appendix}

\end{document}